%% file: main.tex
\documentclass[sn-mathphys-num]{sn-jnl}

\usepackage{graphicx}%
\usepackage{multirow}%
\usepackage{amsmath,amssymb,amsfonts}%
\usepackage{amsthm}%
\usepackage{mathrsfs}%
\usepackage[title]{appendix}%
\usepackage{xcolor}%
\usepackage{textcomp}%
\usepackage{manyfoot}%
\usepackage{booktabs}%
\usepackage{algorithm}%
\usepackage{algorithmicx}%
\usepackage{algpseudocode}%
\usepackage{listings}%
\usepackage{lineno}

\DeclareMathOperator*{\argmin}{arg\,min}

\usepackage{tabularx}
\usepackage{multirow}
\usepackage{array}

\input{sections/def}

\theoremstyle{thmstyleone}%
\theoremstyle{thmstyletwo}%

\theoremstyle{thmstylethree}%

\begin{document}

\title[Article Title]{Learning Physical Interaction Skills from Human Demonstrations}


\author*[1]{\fnm{Tianyu} \sur{Li}}\email{tli471@gatech.edu}

\author[2]{\fnm{Hengbo} \sur{Ma}} 

\author[1]{\fnm{Sehoon} \sur{Ha}}
\equalcont{Equally Contributed.}

\author[3]{\fnm{Kwonjoon} \sur{Lee}}
\equalcont{Equally Contributed.}


\affil[1]{ \orgname{Georgia Institute of Technology}, \orgaddress{\country{USA}}}

\affil[2]{ \orgname{Work Done at Honda Research Institute, \orgaddress{\country{USA}}}}

\affil[3]{ \orgname{Honda Research Institute}, \orgaddress{\country{USA}}}


\newcommand{\kwon}[1]{\textcolor{teal}{[Kwonjoon: #1 ]}}


\abstract{
    Learning physical interaction skills—such as dancing, handshaking, or sparring—remains a fundamental challenge for agents operating in human environments, particularly when the agent's morphology differs significantly from that of the demonstrator. Existing approaches often rely on handcrafted objectives or morphological similarity, limiting their capacity for generalization. Here, we introduce a framework that enables agents with diverse embodiments to learn whole-body interaction behaviors directly from human demonstrations. The framework extracts a compact, transferable representation of interaction dynamics—called the Embedded Interaction Graph (EIG)—which captures key spatiotemporal relationships between the interacting agents. This graph is then used as an imitation objective to train control policies in physics-based simulations, allowing the agent to generate motions that are both semantically meaningful and physically feasible. We demonstrate BuddyImitation on multiple agents, such as humans, quadrupedal robots with manipulators, or mobile manipulators and various interaction scenarios, including sparring, handshaking, rock-paper-scissors, or dancing. Our results demonstrate a promising path toward coordinated behaviors across morphologically distinct characters via cross-embodiment interaction learning.

}




\maketitle

\input{sections/intro}

\input{sections/results}

\input{sections/discussions}

\input{sections/method}



\bibliography{sn-bibliography}


\end{document}

%% file: sections/def.tex
\newcommand{\charpose}{q}
\newcommand{\refcharpose}{{\charpose}}

\newcommand{\IG}{\Phi}

\newcommand{\interconsistR}{r^{ic}}

%% file: sections/intro.tex
\label{sec: intro}

Physical interactions—such as dancing, handshaking, or competitive sports—are fundamental to daily human life. Mastering these behaviors is essential for robots, particularly non-humanoid ones, to function seamlessly in human environments. Robots may conduct daily activities such as object handover or participate in social rituals. Such interactions are critical not only for effective collaboration but also for fostering trust, acceptance, and intuitive communication between humans and machines. Beyond robotics, the ability to synthesize realistic physical interactions is also central to computer graphics and animation, where the goal is to produce lifelike and engaging character behaviors in films, games, and virtual worlds.

Mastering physical interaction skills presents significant challenges due to the complex and dynamic nature of inter-agent coordination. Such interactions demand synchronized pose transitions, consistent contact handling, and a nuanced understanding of interaction semantics—all of which are difficult to formalize analytically or encode through task-specific objective functions. Consequently, designing reward structures or rule-based policies for interaction is often labor-intensive and fails to generalize across scenarios. An appealing alternative is to acquire these skills through imitation of human demonstrations, which inherently capture the timing and structure of coordinated motions. While prior work has demonstrated the effectiveness of this approach for human-like agents~\citep{zhang2023simulation}, it does not readily extend to non-humanoid embodiments, such as quadrupedal robots with manipulators, where substantial kinematic and morphological differences render direct imitation infeasible.

Intelligent animals often display the ability to imitate human behavior despite substantial morphological differences. For example, dogs can mimic yoga poses by observing their owners, suggesting an inherent capacity to learn skills through imitation across embodiments. This phenomenon motivates the concept of cross-embodiment imitation learning, which seeks to enable agents to acquire new skills by observing demonstrators with differing body structures. Prior work has primarily explored this problem in the context of homomorphic motion retargeting, where the demonstrator and learner share identical or similar skeletal topologies but differ in limb lengths or joint configurations. Early methods formulated motion retargeting as an optimization problem based on extracted motion features~\citep{gleicher1998retargetting, choi2000online, tak2005physically}. More recently, data-driven approaches have gained traction, leveraging large motion capture datasets for supervised~\citep{delhaisse2017transfer, jang2018variational} and unsupervised learning~\citep{villegas2018neural, aberman2020skeleton}. A handful of studies have addressed retargeting between characters with more diverse morphologies~\citep{yamane2010controlling, choi2020nonparametric, aigerman2022neural, rhodin2015generalizing, rhodin2014interactive, seol2013creature, kim2022human, li2023ace}, although these efforts largely remain within the realm of kinematic modeling. Cross-embodiment imitation has also been explored in physically simulated environments using reinforcement learning and latent variable models~\citep{grandia2023doc, feng2023genloco, shafiee2023manyquadrupeds, li2023crossloco, RobotMover2025, bohlinger2024one}. However, these approaches focus exclusively on single-agent tasks and do not address the challenges of multi-agent coordination. In contrast, we consider the more complex problem of learning physical interaction behaviors from human-to-human demonstrations, where agents must interpret the spatiotemporal dynamics between characters—such as relative motion, contact timing, and coordination—and adapt these semantics to their own morphology, which may vary substantially in joint structure, limb configuration, or degrees of freedom. This setting presents two core challenges: how to numerically model the semantics of interaction, and how to adapt those semantics to new agent embodiments.

\begin{figure}[h]
    \label{fig:brief_pipeline}
    \centering
    \includegraphics[width=0.995\textwidth]{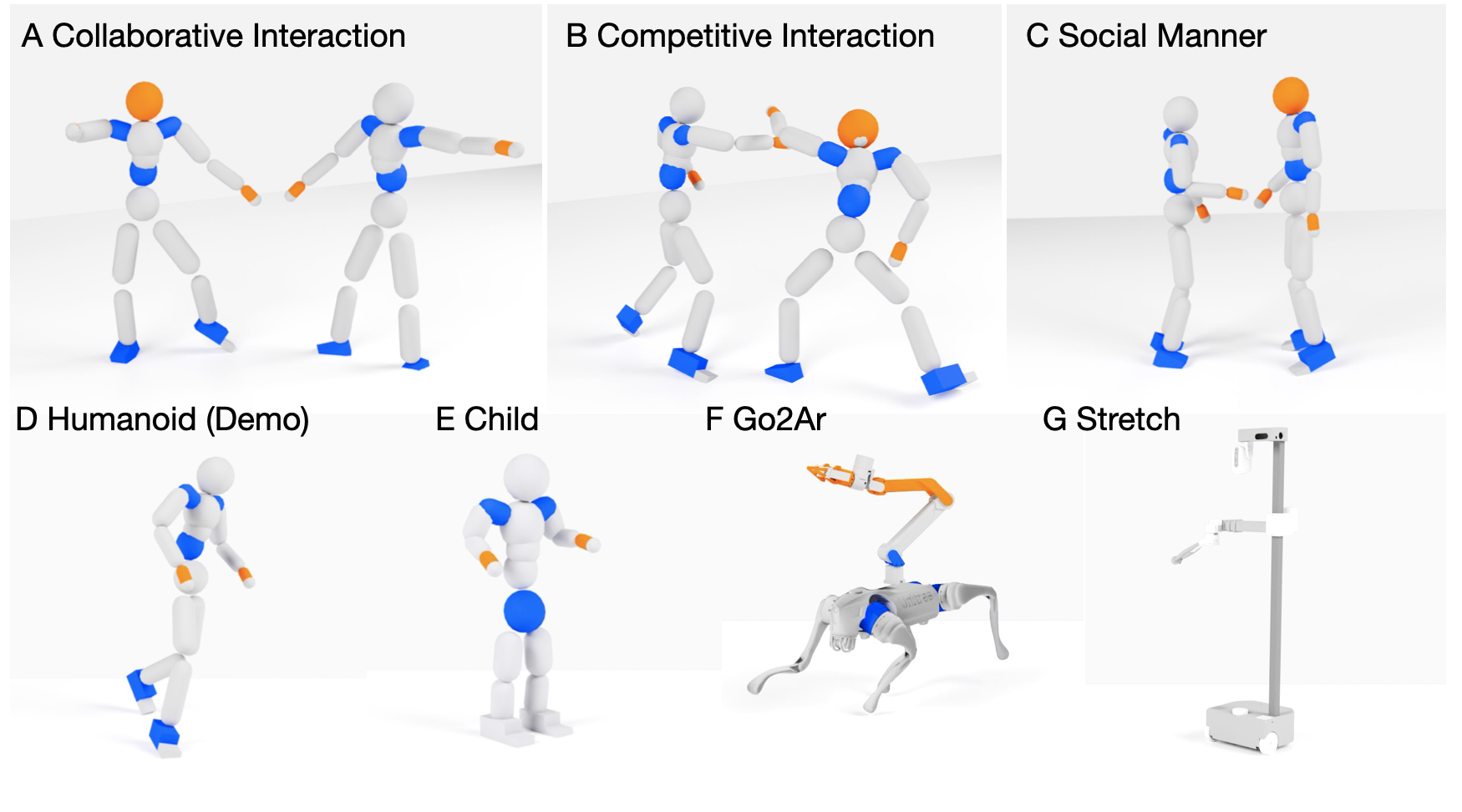}
    \caption{Demonstration interaction types and agent embodiments. This work focuses on physical interactions spanning three categories: (A) collaborative interactions, such as dancing; (B) competitive interactions, such as sparring; and (C) social interactions, such as handshaking. The interactions are transferred to agents with four distinct embodiments: (D) Humanoid, based on the SMPL kinematic structure, consisting of 22 rigid bodies and 21 actuated joints, yielding an action space of $a \in \mathbb{R}^{63}$. (E) Child, which also uses the SMPL structure but has half the size and mass of the Humanoid character. (F) Go2Ar, a legged-manipulation robot built by mounting an Arx5 arm (6 revolute joints, 0.72 m length) onto a Go2 quadruped robot. Each of its four legs has 3 revolute joints, resulting in an action space of $a \in \mathbb{R}^{18}$. (G) Stretch, a mobile manipulation robot with two actuated wheels and a single arm with 4 prismatic joints, for a total action space of $a \in \mathbb{R}^{6}$.}
\end{figure}

Numerical modeling of interaction semantics has been explored through a range of approaches. Early methods rely on expert-crafted constraints~\citep{liu2006composition, kwon2008twoChar, shum2010simulating} or patch-based techniques~\citep{lee2006motion, kim2012tiling, won2014generating, yersin2009crowd} to capture the structure of interactions observed in demonstration data. While effective in controlled settings, these approaches depend heavily on domain-specific heuristics, limiting their generalization across interaction types and agent morphologies. More recently, diffusion models have been employed to implicitly model inter-agent dynamics~\citep{liang2023intergen, shafir2023human}, yielding visually realistic interaction sequences. However, such representations lack interpretability and cannot be directly transferred across embodiments, making them unsuitable for use in cross-morphology imitation. A more general strategy involves the use of interaction descriptors—structured representations that encode spatial and temporal relationships between agents—to quantify semantic differences between original and altered interactions~\citep{al2013relationship, kim2014interactive, kim2021interactive, ho2014multi, ho2010spatial, jin2018aura}. These descriptors have been applied in conjunction with optimization methods such as quadratic programming~\citep{mordatch2012discovery, otani2017adaptive, vaillant2016multi} and reinforcement learning~\citep{won2021control, zhang2023simulation} to generate plausible interaction scenes. Yet, existing descriptors are primarily designed for humanoid agents and lack the flexibility required for generalization to morphologically diverse characters. 
For instance, a na\"ive approach that simply follows an existing descriptor, such as the Interaction Graph~\citep{zhang2023simulation}, may lead to collapsed motions because a full graph does not take different body sizes and structures into consideration.

Beyond modeling interaction semantics, a key challenge lies in adapting these interactions to new agents in a manner that is both natural and physically realistic. Effective adaptation must account for the embodiment of the target agents, ensuring that generated motions are not only visually plausible but also dynamically feasible under physical constraints. The problem of generating physically realistic motion has long been central to both robotics and computer animation. Classical approaches typically formulate control as an optimization problem, employing structured controllers—such as finite state machines—to produce realistic behaviors~\citep{yin2007simbicon, yin2008continuation, coros2010generalized, de2010feature, liu2012terrain, tan2014learning}. More recently, deep reinforcement learning (DRL) has emerged as the dominant paradigm for training motor skills in physics-based environments~\citep{hwangbo2019learning, lee2020learning, Luo2023PerpetualHC, Peng2018deepMimic, Peng2020RoboImitation, Smith2023TWiRL, won2022physics, li2019using, xie2018feedback, Peng2021AMP, Peng2022ASE, Xu2023CompositeAMP}. DRL methods have demonstrated the ability to generate highly dynamic motions, including parkour~\citep{Peng2018deepMimic, hoeller2024anymal}, soccer~\citep{xie2022learningSoccer, liu2022motor, Ji2022A1Shooting, haarnoja2023dmSoccer}, and tennis~\citep{zhang2023vid2player3d}. Hierarchical learning frameworks further improve training efficiency and stability, particularly in tasks requiring long-horizon planning or intricate coordination~\citep{won2020scalable, Peng2022ASE, li2021planning, hoeller2024anymal}.

In this work, we introduce BuddyImitation, a framework that enables agents with diverse morphologies to acquire sophisticated physical interaction skills directly from human demonstrations. The framework proceeds in two stages. First, the Interaction Embedding module encodes the spatiotemporal dynamics of inter-agent interactions into a low-dimensional, sparse graph representation shared across different embodiments. Second, the Interaction Transfer module trains the target agent by leveraging the learned representation as an imitation objective in a physics-based environment, allowing it to reproduce interaction behaviors that are both physically feasible and semantically consistent with the original demonstration. We evaluate BuddyImitation across a wide range of agents and interaction scenarios, including collaborative behaviors such as dancing, competitive activities such as sparring, and socially grounded gestures such as handshaking. Some demonstration examples and an overview of the involved embodiments are shown in Figure:~\ref{fig:brief_pipeline}. Our experiments show that the proposed framework enables diverse agents to internalize and reproduce interaction semantics while adapting to their own physical embodiment. Extensive ablations further validate the importance of the learned interaction representation and the transfer mechanism, highlighting the flexibility and robustness of our approach to cross-embodiment interaction learning.

%% file: sections/results.tex
\section{Results}
\label{result}

We conducted all experiments on a PC equipped with an AMD Ryzen Threadripper PRO 5975WX CPU and a NVIDIA 3080 GPU. We use the Genesis environment~\citep{Genesis} as the physics-based simulation backend. Human interaction demonstrations are sourced from the InterHuman dataset~\citep{liang2023intergen}, a large-scale 3D motion dataset capturing diverse physical interactions between two people. From this dataset, we select 25 interaction scenes covering a wide range of semantically distinct behaviors. Each demonstration lasts between 10 and 20 seconds. Our experiments involve four simulated agents with varied embodiments, illustrated in Figure~\ref{fig:brief_pipeline}, to evaluate the framework’s ability to generalize across morphologies.

\begin{figure}[h!]
    \label{fig:result_overview}
    \centering
    \includegraphics[width=0.99\textwidth]{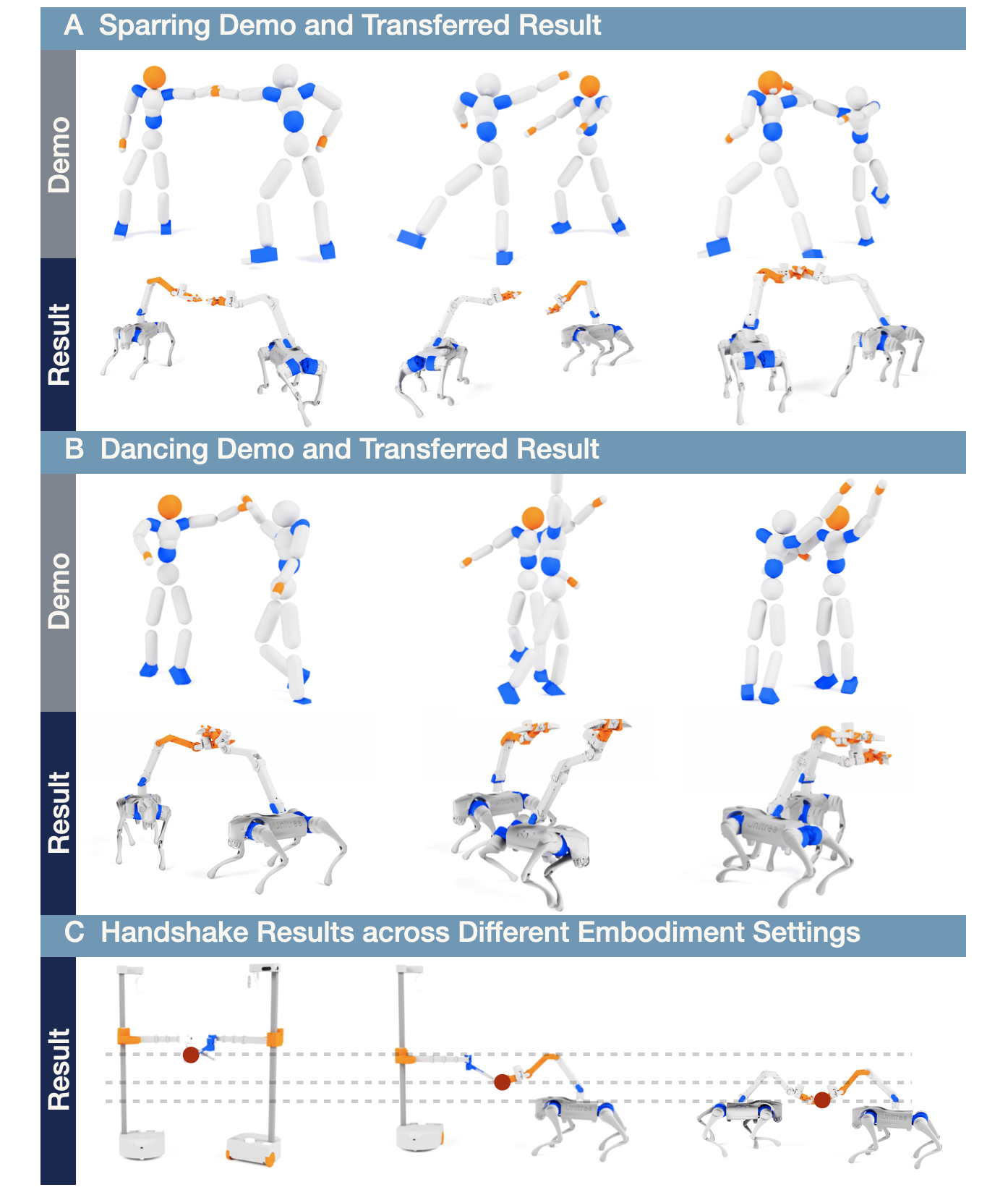}
    \caption{Overview of interaction transfer results. (A) Sparring: Two Go2Ar robots, which differ significantly in morphology from the human demonstrators, successfully learn the sparring activity. The robots capture key interaction patterns such as striking and dodging. (B) Dancing: Go2Ar robots learn a dancing sequence, reproducing motion patterns such as arm reaching, raising, and circular movement. (C) Handshaking: The handshaking behavior is learned across different character pairings. The interaction adjusts based on the embodiment of each agent. For example, two Stretch robots shake hands at a higher position, while a Stretch robot lowers its arm when shaking hands with a Go2Ar. Two Go2Ar robots perform the handshake at the lowest position, reflecting their shorter limb structure.
    }
\end{figure}

The results show that the proposed framework enables agents with diverse embodiments to learn human-like physical interaction behaviors by imitating human demonstrations. For example, the Go2Ar legged manipulator successfully acquires highly dynamic behaviors such as sparring and dancing, executing interaction patterns using only a single manipulator. The Stretch robot learns to extend its prismatic-jointed arm and shake its gripper to replicate handshaking gestures in social interactions. Representative examples of the learned behaviors, alongside the original human demonstrations, are shown in Figure~\ref{fig:result_overview}. For a more comprehensive view of the interaction quality and dynamics, we refer readers to the supplementary video.

Our results indicate that the proposed method effectively captures the inter-agent dynamics present in the demonstrations while adapting the movements to the embodiment of each agent. In the sparring scenario, for example, the human demonstrators alternate between offense and defense motions—using their arms to ``attack'' the opponent, shifting their body to ``dodge'' incoming strikes, and positioning their limbs to block contact. The Go2Ar quadrupedal robots successfully reproduce these interaction patterns, demonstrating coordinated ``attack,'' ``dodge,'' and ``defend'' behaviors. Despite being equipped with only a single manipulator, each robot automatically learns to mimic the active arm behaviors observed in the demonstrations. Furthermore, because of their quadrupedal structure and limited range of limb motion, the robots naturally adapt human leg movements into body shifts suited for four-legged locomotion. This results in similar root trajectories between the human and robot motions, but with distinct leg movement strategies adapted to the robots’ morphology.

We observe that the proposed framework can lead to the result that agents able to adjust behavior based on the morphology of the collaborating agent. In Figure~\ref{fig:result_overview}c, we show how a human handshaking demonstration is transferred to three different character pairings: Go2Ar vs. Go2Ar, Go2Ar vs. Stretch, and Stretch vs. Stretch. In each setting, the characters adjust the height of the handshake according to their respective embodiments. The Go2Ar robot, being relatively small with short legs, performs handshakes at a lower height when paired with another Go2Ar. When interacting with the taller Stretch robot, Go2Ar raises its arm to reach the partner’s gripper, while Stretch lowers its arm to reciprocate. In Stretch vs. Stretch, both agents meet at a higher handshake position consistent with their morphology. These results suggest that the proposed framework enables agents to adaptively adjust their interaction strategies in response to the embodiment of their collaborators—an essential capability for real-world human-robot interaction.

The results also demonstrate that the proposed method is robust to imperfections in the demonstration data. In some cases, the human demonstrations include physically infeasible motions—such as body parts penetrating the ground or a demonstrator's arm intersecting with their collaborator due to miscellaneous errors during motion capture. Because the agent’s control policy is trained in a physics-based simulation environment, these infeasible motions result in premature episode termination and correspondingly low rewards. As a result, the agent learns to avoid such behaviors in order to maximize its cumulative reward. This finding highlights the robustness of our framework in learning from noisy or imperfect demonstration sources, such as motion capture datasets.

In the following subsections, we provide additional analysis and empirical studies of the proposed approach. We begin by examining the Interaction Embedding module, which learn the inner dynamic of the interactions. We then evaluate the Interaction Transfer module, which enables new agents to reproduce the learned interactions using guidance from Interaction Embedding module.

\subsection{Interaction Embedding Module for Interaction Modeling}

\begin{figure}[h!]
\centering
\includegraphics[width=0.9\textwidth]{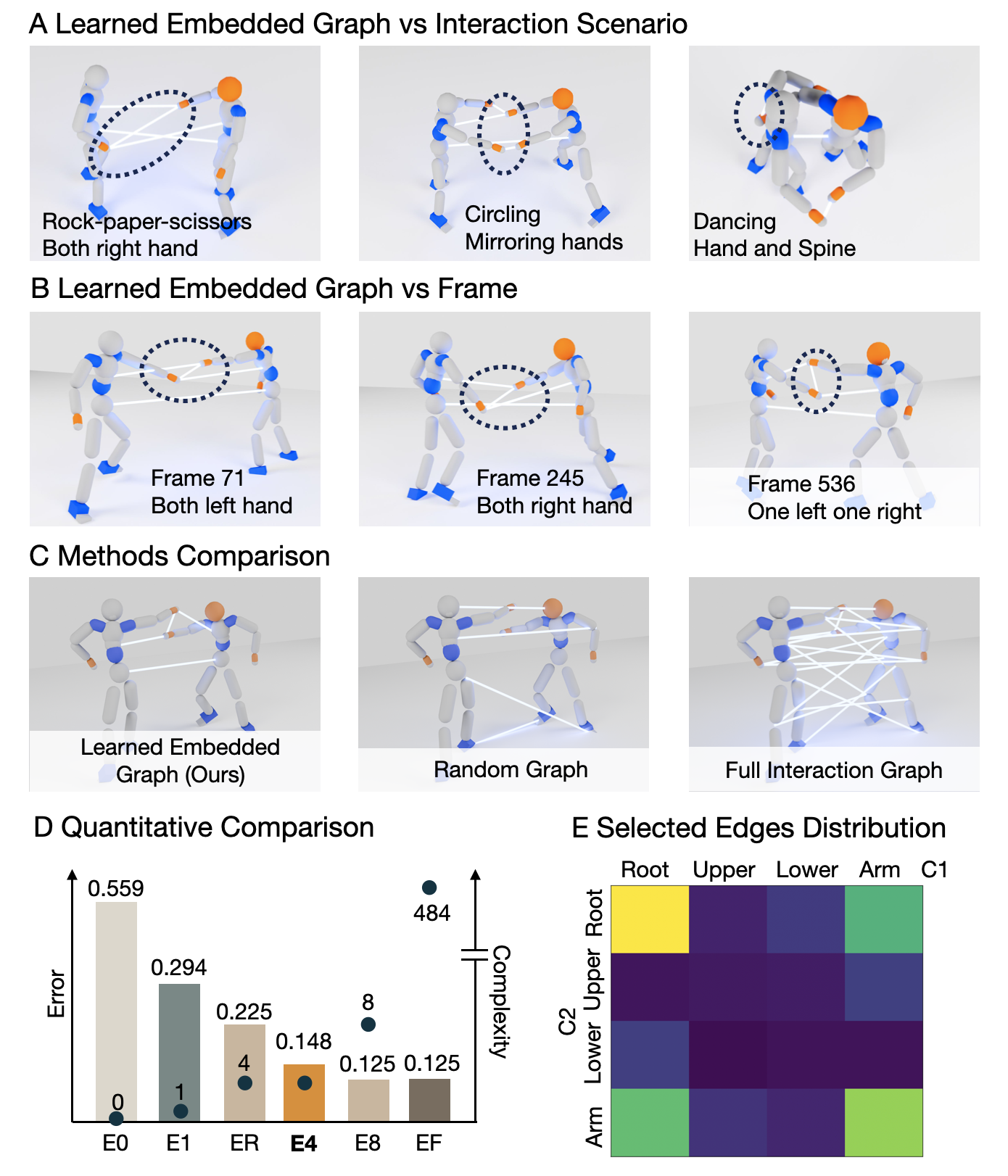}
\caption{Overview of Embedded Interaction Graph learning results.
(A) The learned embedded graph varies across interaction scenarios. For example, in Rock-Paper-Scissors (RPS), it focuses on the movement of both characters’ right hands, while in other activities it may emphasize mirrored hand positions or the relationship between the hand and spinal joints. (B) The embedded graph also evolves over time. In Sparring, the graph dynamically adapts to the currently dominant hand of each character.(C) Visual comparison between the learned embedded graph, a randomly sampled graph with the same number of edges, and the full interaction graph. (D) Quantitative comparison of pose reconstruction accuracy across different graph configurations. (E) Heatmap visualization of spatial edge selection frequency in the learned embedded graph. To generate the aggregated heatmap, we divide the human body into four regions: Root (core, spine), Upper (chest, neck, head), Lower (legs), and Arms. Lighter colors indicate higher edge selection frequency. The results show that the most influential relationships are typically between the root and arm regions of both characters, suggesting that a character’s motion is primarily influenced by their partner’s arm and core movements.}
\label{fig:emb_results}
\end{figure}

The Interaction Embedding module learns to represent inter-agent dynamics using sparse graphs derived from human demonstrations. These embedded graphs capture the essential components of the interaction by selecting a subset of joints on agents that are most predictive of the agents' future movements. Examples of learned embedded graphs across various interaction scenarios are shown in Figure~\ref{fig:emb_results}. The evolution of the graph edges highlights the key relational features that drive coordinated behavior between agents. In this subsection, we analyze the properties of the learned embedded graphs and evaluate their effectiveness in modeling interaction semantics.

The results show that the learned embedded graph dynamically adapts based on the type of activity and the poses of the interacting characters. In the Handshaking and Rock-Paper-Scissors scenarios, the graph consistently focuses on the relative root positions and the poses of the characters’ right arms. In the Sparring activity, the structure of the graph varies according to the spatial configuration and motion of the agents. In some frames, it emphasizes the relative movement of the right arms; in others, it shifts focus to the root positions and left arm poses. For the Circling activity, where two characters hold hands and move in a circular motion, the graph captures the mirrored hand positions between the agents. These observations suggest that the learned embedded graph is both selective and adaptive, highlighting task-relevant interaction cues in a dynamic and context-dependent manner.

The learned embedded graph provides a compact yet interpretable representation of the interaction. For example, in the Sparring scenario, when the graph emphasizes the right hand of one character and the left hand of the other, it highlights a moment in which the semantic focus is on those limbs—corresponding tod an attack and defense pattern. In the Handshaking scenario, the graph consistently selects edges connecting the characters’ right hands and their relative root positions, capturing the essential spatial alignment required for a successful handshake. These examples illustrate that the embedded graph is not only functionally predictive but also semantically meaningful, offering an interpretable abstraction of the underlying interaction dynamics.

\subsubsection{Quantitative Analysis}
We quantitatively evaluate the proposed interaction modeling method by comparing it against several baseline approaches in the task of predicting the future pose of the demonstrator. A more effective model should yield higher prediction accuracy. Performance is assessed by rolling out 120 frames (equivalent to 2 seconds) from a randomly selected starting point within an interaction sequence. Each evaluation is averaged over 500 trials to ensure statistical reliability. Prediction error is computed as the mean squared error (MSE) between the predicted and ground-truth poses. A summary of the results is presented in Figure~\ref{fig:emb_results}d.

In the first experiment, we evaluate the effectiveness of our learning-based edge selection algorithm by comparing it to a randomly generated graph(\textbf{ER}). The random graph is constructed by selecting the same number of edges as the learned embedded graph, but without applying any learning-based selection criteria. Both models are trained using the same motion prediction setup, differing only in the graph structure used as input. The results show that the learned embedded graph yields significantly more accurate future pose predictions, closely matching the ground-truth motion, while the random graph fails to reconstruct the interaction dynamics. As shown in Figure~\ref{fig:emb_results}d, the learned graph achieves a substantially lower prediction error than the random baseline, indicating its superior ability to capture the essential structure of inter-agent interactions.

We conduct an ablation study to evaluate the role of edge sparsity in the embedded interaction graph. Specifically, we investigate how the number of selected edges affects motion prediction accuracy, using mean squared prediction error as the evaluation metric. We compare five settings: a 4-edge embedded graph \textbf{E4}, our default), no graph \textbf{E0}, a single-edge graph \textbf{E1}, an 8-edge graph \textbf{E8}, and a fully connected graph with 484 edges \textbf{EF}.  Overall, prediction error decreases as more informative edges are included. Without any interaction information from the partner \textbf{E0}, the character is unable to model the interaction and produces erratic behaviors, leading to the highest prediction error. Introducing a single edge \textbf{E1} already yields a substantial reduction in error—by 0.265—indicating that even minimal partner information is beneficial. Interestingly, the selected edge in E-1 often corresponds to the shortest distance between keypoints on the two characters, highlighting the model’s preference for spatially relevant features. Both E-4 and E-8 achieve high prediction accuracy, while the fully connected graph \textbf{EF} performs worse due to overfitting and the difficulty of processing overly dense input structures. These findings underscore the importance of edge sparsity for both predictive accuracy and training efficiency.

The learned interaction graph, together with its associated state trajectory, is used as an interaction consistency reward for training new characters to replicate the demonstrated interaction behavior. In subsequent experiments, we adopt the graph configuration from the \textbf{E4} setting, as it offers a favorable balance between accurate interaction modeling and efficient data utilization.

\subsection{Transferring Interaction to New Characters}

\begin{figure}[h!]

\centering
\includegraphics[width=0.95\textwidth]{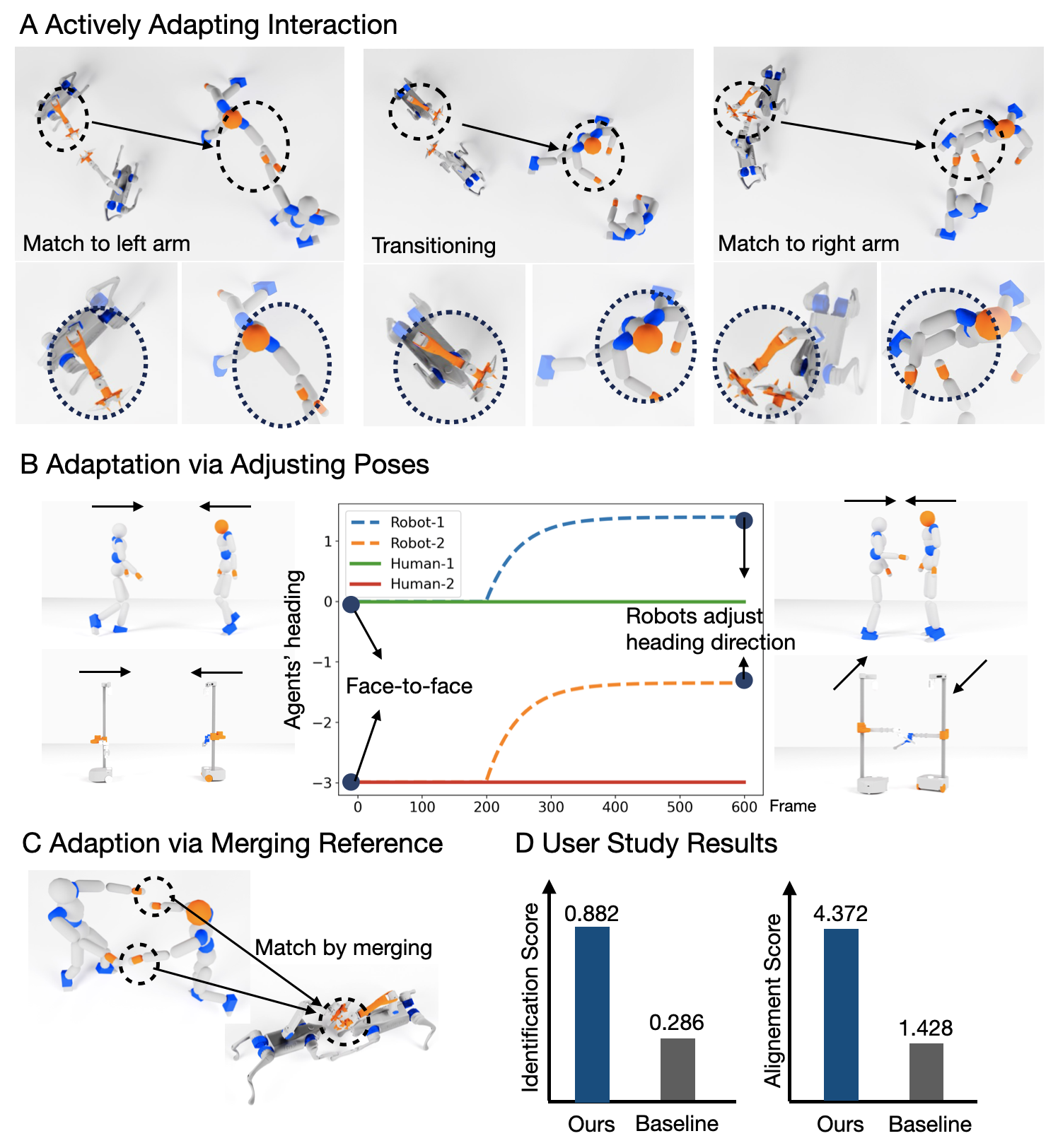}
\caption{(A) A single-arm robot dynamically switches its referenced demonstrator arm from left to right based on the dominant arm in each interaction state. (B) Unlike human demonstrators who perform handshakes face-to-face, Stretch robots learn to rotate their bodies to shake hands, compensating for their prismatic joint structure. This demonstrates embodiment adaptation through pose adjustment to maintain semantic alignment. (C) Robots with a single arm learn to approximate bimanual motions by merging references from both demonstrators’ arms, showing that they can preserve interaction semantics despite morphological limitations. (D) 
We compare our method against a baseline to assess whether users can correctly identify the interaction activity from motion videos without access to the original demonstrations, and to evaluate the perceived alignment between the learned behaviors and demonstration videos. The results show that our method significantly outperforms the baseline in both activity recognition and semantic consistency.}
\label{fig:transfer_results}
\end{figure}

The Interaction Transfer module enables the transfer of learned interaction behaviors to characters with significantly different morphologies. By leveraging the embedded interaction graph generated by the Interaction Embedding module, this component trains control policies that encourage new agents to produce interaction behaviors aligned with those observed in the demonstrations. In this section, we evaluate the module's effectiveness in both preserving interaction semantics and adapting behavior to new embodiments. Some results are shown in Figure~\ref{fig:transfer_results}.

According to our results, we find that even characters with a single manipulator, such as Go2Ar, can successfully learn interaction behaviors that originally involve bimanual motion—such as in \textit{Sparring}. This is made possible by the learned embedded interaction graph, which captures the essential semantic role of each limb throughout the interaction. As a result, Go2Ar learns to switch its imitation reference limb and only focus on the limb motion that dominate the current status of the interaction. This leads to alternate between imitating demonstrator's left and right arm as presented in Figure~\ref{fig:transfer_results}a.

We observe that the learned control policies adapt movements to the physical constraints of new character embodiments. For example, smaller characters tend to follow root trajectories that stay closer to the demonstrators', compensating for limited reach or speed. The Go2Ar robot adapts to fast-turning behaviors in the demonstration by increasing its stepping frequency. The Stretch robot, which relies on a prismatic joint for arm extension, rotates its base to reach interaction points—unlike the human demonstrators, who use a ball-jointed shoulder as shown in Figure~\ref{fig:transfer_results}b. In the \textit{Circling} activity, where two humans hold hands and move in a circular path, Go2Ar adapts by using its single manipulator to simulate hand-holding, while Stretch uses its gripper. These examples illustrate the framework’s ability to generalize semantic interaction patterns across morphologically diverse characters. However, some limitations remain. For instance, sharp turning motions in the demonstration cannot be fully replicated by Go2Ar due to its limited turning agility as a quadruped. Such cases highlight the inherent challenges of transferring interactions to agents with restricted actuation or mobility.

We conducted a user study to quantitatively evaluate the performance of our interaction learning framework. The study consisted of two parts. In the first part, participants were shown a series of learned interaction results without access to the corresponding demonstration videos and were asked to identify the interaction scenarios. Performance was measured by the identification accuracy. In the second part, participants were shown both the learned interaction results and the corresponding demonstration videos and were asked to rate the semantic alignment on a scale from 0 to 5, with higher scores indicating better alignment. For comparison, we included results from an inverse kinematics-based motion retargeting baseline. We recruited 50 participants with diverse backgrounds, including robotics, computer animation, human–robot interaction, as well as participants with no relevant technical background. A summary of the results is presented in Figure~\ref{fig:transfer_results}. The findings indicate that users were better able to identify the interactions produced by our method, and they rated our results as more closely aligned with the original demonstrations—suggesting stronger semantic preservation.

%% file: sections/discussions.tex
\section{Discussion}\label{discussion}

In this work, we address the challenge of enabling characters with diverse morphologies to learn physical interaction behaviors directly from human demonstrations. Our approach first learns a sparse graph representation that captures the inter-agent motion dynamics, and then uses this learned graph as an imitation reference for training new agent policies in simulation via reinforcement learning. The results demonstrate that the proposed framework allows characters with non-humanoid morphologies to acquire complex and meaningful interaction skills, including collaborative, competitive, and social interactions. There are several key aspects of this approach and its implications that merit further discussion.

The learned embedded interaction graphs efficiently capture the essential components of interactive motion, enabling accurate prediction of future interaction dynamics. However, the experiments in this work primarily focus on two-character interactions due to limitations in available datasets. In the future, we aim to extend this interaction modeling approach to more complex scenarios, including interactions among larger groups of characters or between characters and objects. Beyond interaction transfer, we see potential for applying this method in other domains such as sports analysis or social behavior understanding. Because the embedded graph explicitly highlights key relationships between characters, it could help reveal underlying motivations and intentions in interactions, offering deeper insights into social dynamics, improving human–robot collaboration, and enhancing the realism of virtual characters in gaming and simulation.

Secondly, when transferring interactions to new character settings, we treat each configuration independently. This is because each character has its own distinctive sensory and actuation spaces, requiring a dedicated policy that takes both agents’ states as input to imitate a given interaction demonstration. Additionally, for different interaction scenarios—even with the same characters—we train separate policies. As a next step, we aim to develop a unified and generalizable control policy capable of handling multiple interaction types and character settings within a single framework. Achieving this goal could involve designing a shared observation and action space that encompasses diverse embodiments, enabling more flexible and robust interaction learning. We also envision incorporating interaction-semantic conditioning to allow the unified policy to adapt its behavior dynamically when performing different interaction activities.

While our framework shows that complex human interactions can be transferred to characters with diverse morphologies, an important future direction is to move beyond direct motion imitation and toward acquiring interaction skills that incorporate deeper reasoning, adaptability, and generalization. For example, in competitive scenarios like sparring, an agent should not only reproduce offensive and defensive movements but also develop strategies to adapt to an opponent’s behavior and refine its performance over time. Bridging this gap will require new methods that integrate semantic understanding and high-level reasoning with robust low-level motion control, combining advances in task and motion planning with reinforcement learning. Ultimately, progress in this direction could enable agents to interpret, adapt, and generate novel interactions autonomously, moving imitation learning closer to the broader goal of embodied artificial general intelligence.


%% file: sections/method.tex
\section{Methods}
\label{sec:method}

\begin{figure}[h!]
    
    \centering
    \includegraphics[width=0.95\textwidth]{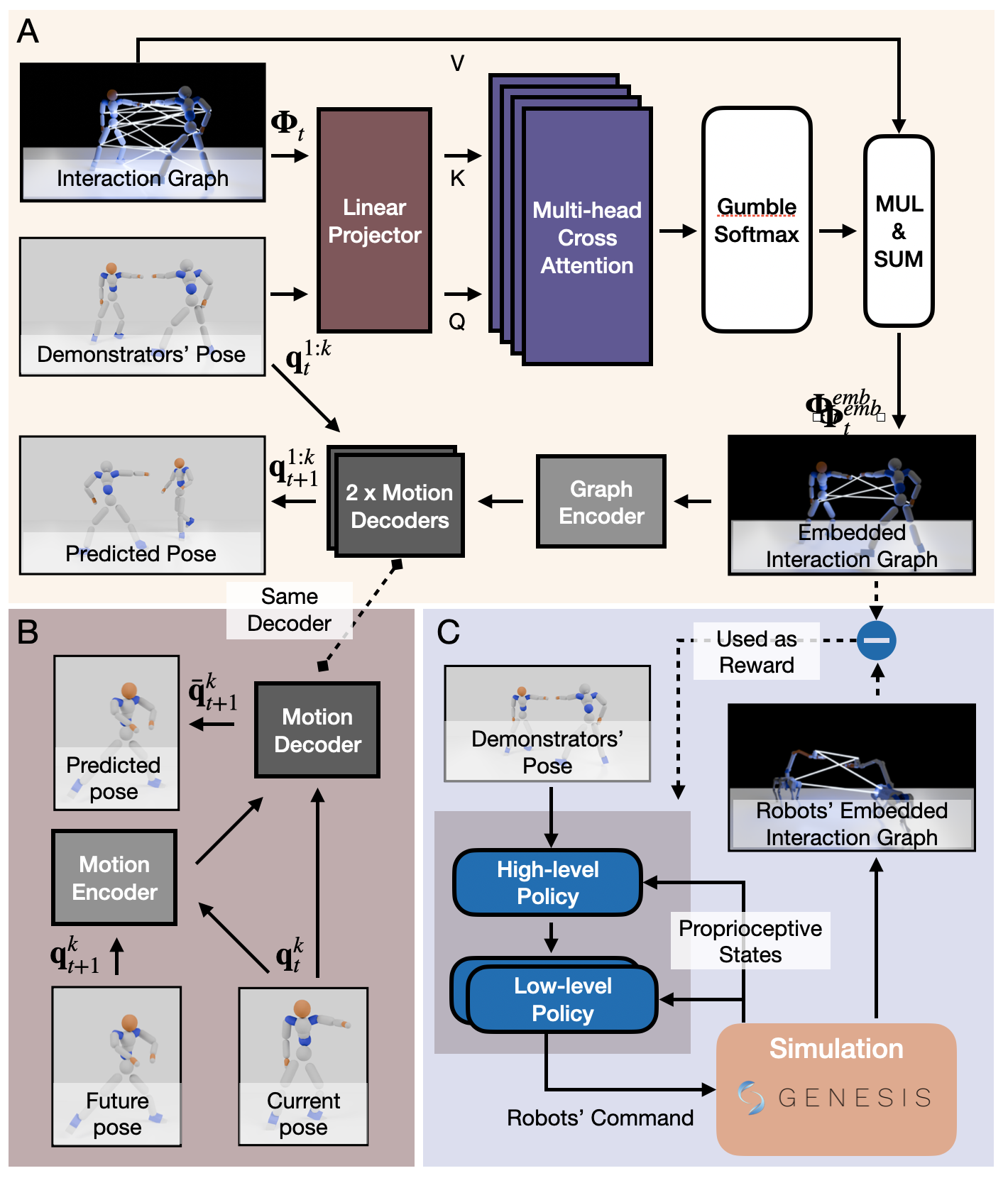}
    \caption{Overview of the proposed method. (A) Interaction Embedding Module: A multi-head cross-attention mechanism selects the most informative edges from the fully connected interaction graph, producing a sparse embedded interaction graph. This embedded graph is used to predict each demonstrator’s future pose using a pretrained single-character motion decoder. (B) Pretrained Motion Decoder: A motion variational autoencoder (VAE) is trained to capture diverse pose transitions. Only the decoder is retained for inference after training. (C) Interaction Transfer Module: This module transfers the demonstrated interaction to new agents by aligning the embedded interaction graph through reinforcement learning in a physics-based simulation. A hierarchical control policy is used, consisting of character-specific low-level policies governed by a shared high-level policy. Low-level controllers are initialized via pretraining and further optimized during interaction learning.}
    \label{fig:method_overview}
\end{figure}

Our proposed framework, BuddyImitation, is an imitation learning framework designed to enable diverse agents to acquire physical interaction skills from human demonstrations. The framework comprises two key learning components: the Interaction Embedding module and the Interaction Transfer module, which correspond to the two main stages of the pipeline. The Interaction Embedding module learns an explicit, low-dimensional representation that numerically encodes the semantics of interaction. The Interaction Transfer module then uses this representation to transfer the demonstrated behavior to new agents by aligning the interaction semantics with their embodiment. An overview of the BuddyImitation framework is illustrated in Figure~\ref{fig:method_overview}. In the following subsections, we begin by introducing the necessary background, followed by detailed descriptions of the Interaction Embedding and Interaction Transfer modules.

\subsection{Preliminary}
\subsubsection{Interaction Graph}
The Interaction Graph~\citep{zhang2023simulation} is a spatial graph structure that numerically encodes the semantics of inter-agent interactions. The graph $G(V,E)$ consists of a set of nodes $V$ and edges $E$. Nodes are placed on salient points of every character, such as revolute joints, and edges are defined between pairs of nodes between characters. Each edge is assigned a $6$D edge feature vector $\phi^{ij} \in \mathbb{R}^6$ that captures the relative position and midpoint of the connected joints in the world frame, where $i$ and $j$ represent the joint indices in the corresponding character. The feature of the Interaction Graph is defined by the collective state of edges $\IG_t = [\phi^{11}_t, \phi^{12}_t, \cdots \phi^{mn}_t ]$, where $m$ and $n$ are the number of nodes. This feature $\IG_t$ can be derived from the interaction graph $G$ and the joint positions of the characters $\refcharpose_t = [\refcharpose^1_t, \cdots, \refcharpose^K_t]$ where a superscript refers to the index of the character, where $K$ is the number of characters. Because of the kinematic structure, all the characters' joint positions, including the opponent, can also be reconstructed from the given one character's pose and the graph feature $\IG_t$:
\begin{align}
\label{eqn:IG_eqv}
\refcharpose^{1}_t, \cdots, \refcharpose^{K}_t, = IK(\refcharpose^k_t, \IG_t).
\end{align}

Our BuddyImitation framework is built upon the concept of the Interaction Graph to model the dynamics of demonstration interactions. However, due to its high dimensionality and complexity, the original Interaction Graph is not well suited for cross-embodiment imitation, where a more compact and transferable representation is needed.

\subsubsection{Reinforcement Learning}
Reinforcement learning (RL) is a widely used approach for enabling agents to acquire new skills through interaction with an environment. At each timestep, an agent receives an observation $o_t$ and selects an action $a_t$ based on its policy $\pi(a_t \mid o_t)$. The selected action is executed in the environment, resulting in a new state $s_{t+1}$ and a scalar reward $r_t = r(s_t, a_t)$. The objective of reinforcement learning is to train a policy $\pi$ that maximizes the expected cumulative discounted reward over time:
\begin{align}
J(\pi) = \mathbb{E}_{p(\tau|\pi)}\left[\sum_{t=0}^{T-1} \gamma^t r_t \right],
\end{align}
where $\tau$ denotes a trajectory generated by executing $\pi$, $T$ is the trajectory horizon, and $\gamma \in [0,1)$ is the discount factor.

In this work, we use reinforcement learning to train agents to perform physical interaction behaviors. Unlike prior approaches that focus on single-agent scenarios, our setting involves simultaneous multi-agent policy learning, where the agents must coordinate their actions to maintain consistent and physically plausible interactions.

\subsection{Interaction Embedding Module}

The objective of the Interaction Embedding module is to encode complex interaction movements into an explicit, low-dimensional numerical representation that captures the most important semantic of the interaction. This representation can then be used as a reference to guide policy learning for agents with different morphologies. To model the interaction numerically, we adopt the concept of the Interaction Graph, where an interaction sequence is represented as a sequence of graph features, denoted $[\IG_1, \cdots \IG_T]$. However, the original Interaction Graph $G$ is fully connected—linking all joints between characters—which results in a large number of edges, many of which are redundant or irrelevant to the core interaction. For example, in a handshaking scenario, the interaction is primarily governed by the motion of the hands and arms, while the legs play a less critical role. Motivated by this observation, the Interaction Embedding module introduces a procedure to learn sparse embedded interaction graphs $G^{\text{emb}}_t(V^{\text{emb}}_t, E^{\text{emb}}_t)$ that preserve essential dynamics while discarding irrelevant information and yield concise feature trajectories: $[\IG^{\text{emb}}_1, \cdots \IG^{\text{emb}}_T]$. These sparse graphs serve as efficient, interpretable surrogates for the original fully connected graphs.
Also note that this embedding graph takes a time index $t$ as input because the semantically important edge can vary over motion segments.
In the following subsections, we first formulate the graph embedding problem, then describe the learning architecture and training procedure.

\subsubsection{Problem Formulation}

Given an interaction trajectory from a pre-collected two human interaction dataset ($K=2$), we denote the 3D joint positions $q$ of all demonstrators as $Q = \{(\refcharpose^{1}_{1},   \refcharpose^{2}_{1}), \cdots, (\refcharpose^{1}_{T},\refcharpose^{2}_{T})\}$, where a superscript indicates the character index and $T$ is the length of the sequence. For simplicity, we denote $\refcharpose_{t} = [\refcharpose^{1}_{t}, \refcharpose^{2}_{t}]$. Because we have access to all joint positions, we can construct a fully connected interaction graph feature $\IG_t$ at each time step $t$ using kinematic information. The goal of the Interaction Embedding module is to learn a sparse embedded interaction graph $G^{emb}_t \subseteq G$ that effectively replaces the full graph while preserving the interaction dynamics.

We model the interaction prediction problem as a Markov Decision Process (MDP), where the future state of the characters depends only on the current state:
\begin{align}
P(\refcharpose_{t+1}|\refcharpose_{1},\cdots,\refcharpose_{T}) = P(\refcharpose_{t+1}|\refcharpose_{t}).
\end{align}
According to the kinematic equivalency as show in Equation~\ref{eqn:IG_eqv}, for a $k$th individual agent we have:
\begin{align}
P(\refcharpose^{k}_{t+1}|\refcharpose_{t}) = P(\refcharpose^{k}_{t+1}|\refcharpose^{k}_{t}, \IG_t).
\end{align}

Our objective is to learn a time-dependent sparse graph $G^{\text{emb}}_t$ and the associated feature $\IG^{\text{emb}}_t$ that can substitute $\IG_t$ in modeling the interactions. The learning objective is therefore to minimize the distributional difference between the predictive distribution conditioned on the full graph $P(\refcharpose^{k}_{t+1} |\refcharpose^{k}_{t}, \IG_t)$ and that conditioned on the learned sparse graph $P(\refcharpose^{k}_{t+1} | \refcharpose^{k}_{t}, \IG^{emb}_t)$. Here, 
the learning objective is to minimize the mean squared error between the predicted and true future poses:
\begin{align}
\argmin_{G} \  L_{recon} = \sum_{k=1}^{K} \ [  \refcharpose^{k}_{t+1} - \Psi(\refcharpose^{k}_{t}, \IG^{emb}_t)  ]^2, \\
\textit{s.t.} \ \IG^{emb}_t = \xi(\refcharpose^{1:K}_{t}, \IG_t), \quad \mathbb{N}(\IG^{emb}_t) = \bar{n}.
\end{align}
Here, $\Psi$ is a trainable pose prediction function which predict the agent's future joint positions given the embedded interaction graph and it current joint positions.
$\xi$ denotes the learned embedding function that selects informative edge features from the full interaction graph, $\mathbb{N}(\IG^{emb}_t) = \bar{n}$ indicates the total number of edge in the learned embedded graph is equal to $\bar{n}$.

\subsubsection{Learning Architecture}

The learning architecture of the \textit{Interaction Embedding module} is illustrated in Figure~\ref{fig:method_overview}. It consists of two main stages: Edge Selection and Pose Prediction. The Edge Selection stage identifies the most informative edges from the full interaction graph $\IG_t$ to construct a sparse graph $G$. The Pose Prediction stage then uses the selected edges to forecast future character poses, which are used to compute the training loss. Both stages are trained jointly in an end-to-end fashion.

In the Edge Selection stage, we employ a multi-head cross-attention mechanism to identify critical edges in the full interaction graph $G$. Specifically, attention is computed between the current pose of a character, $\refcharpose^{k}_t$, and the current graph feature in $\IG_t$. The query($Q$) is the encoded pose of the character, while the keys($K$) are edge embeddings from the full graph, with each edge encoded independently into a latent space. The values correspond to the original (unencoded) edge features.To enforce sparsity, we apply \textit{hard attention} in each attention head, selecting a single edge per head. This results in a filtered interaction graph $G^{\text{emb}}_t$ composed of $k$ edges, where $k$ equals the number of attention heads. The output of the attention mechanism is computed by multiplying the selected edges with the corresponding value vectors, yielding the final embedded graph feature $\IG^{emb}_t$. 

In the Pose Prediction stage, the sparse interaction graph feature $\IG^{emb}_t$ is used to predict the future poses of all characters. It is crucial to note that this generative model predicts the pose of a \textbf{single} human, rather than the poses of all characters in the scene. We employ the same model for each humanoid character within the scene to predict the overall future interaction state. First, $\IG^{emb}_t$ is encoded into a latent vector using a graph encoder. This vector, along with the current pose $\refcharpose^{k}_t$ of the character, is input into a neural pose decoder that predicts the character's future pose $\refcharpose^{k}_{t+1}$. This procedure is repeated for each character, enabling simultaneous prediction of future poses across all agents in the interaction.

\subsubsection{Pretrain Human Motion Decoder}

To support the learning of the embedded interaction graph, we pretrain a single human motion decoder used in the pose prediction stage. The goal of this decoder is to capture the diversity of human pose transitions and accurately reconstruct future poses from the current pose and the embedded interaction graph. This model serves as the backbone of the Interaction Embedding module.

Our motion decoder is based on the Motion Variational Autoencoder (MVAE) framework proposed by \citet{ling2020character}. The MVAE takes the character’s current pose and a latent variable $z_t$, which encodes plausible transitions, to reconstruct the next pose. The latent space is regularized to follow a normal distribution. During inference, only the decoder is used: given the current pose and a sampled $z_t$, it predicts the next pose, enabling autoregressive generation of motion sequences.

We adopt the architecture and training scheme of the MVAE, incorporating several key techniques to improve performance. Scheduled sampling~\citep{ling2020character, zhang2023vid2player3d} is used to stabilize autoregressive predictions. The KL divergence weight $\beta$ is tuned to balance latent space usage and motion fidelity. A large $\beta$ causes the model to ignore $z_t$, leading to overfitting, while a small $\beta$ may result in artifacts such as foot sliding or jitter. We use $\beta = 0.3$ to strike a balance between generalization and motion quality.

\subsubsection{Graph Consistency Loss}

Beside the pose reconstruction loss, we also include a graph consistency loss to ensure that the learned embedded graph remains stable and consistent throughout the interaction trajectory. This is particularly important for maintaining temporal coherence in the interaction dynamics, as well as reducing redundancy in the learned graph structure. Here we use the variance of the outputs from the attention mechanism used in the Edge Selection stage as the graph consistency loss. Specifically, we compute the variance of the Gumbel-Softmax outputs of the attention heads, which reflects how consistently each edge is selected across different time steps. Therefore, the total loss function for the Interaction Embedding module can be expressed as:
\begin{align}
L_{emb} = L_{recon} + \lambda L_{var},    
\end{align}  
where $L_{recon}$ is the reconstruction loss for the predicted poses, $L_{var}=Var(\IG^{emb})$ is the graph consistency loss based on the variance of the attention outputs, and $\lambda$ is a hyperparameter that balances the two losses.

\subsection{Interaction Transfer Module}

The Interaction Transfer module is tasked with transferring interaction movement demonstrations to new character settings while maintaining interaction consistency. Our approach is inspired by the unsupervised reinforcement learning framework~\citep{li2023crossloco}, which formulate a cross-domain-transfer-specialized reward and use reinforcement learning to simultaneously learn new character movements and their correspondence to human motions. The principal challenge here is establishing the correspondence between demonstrated human movements and those of new characters. Our interaction consistency comes from aligning the learned embedded interaction graph between the demonstration and the new character. The interaction consistency reward is designed to guide the policy toward interaction semantic preservation by measuring the difference between embedded graph features of target characters $\hat{\IG}^{emb}_t$, which is a function of the embedding graph $\hat{G}^{\text{emb}}_t$ and the current states $\hat{q}^{1}_t, \hat{q}^{2}_t,$ and reference $\IG^{emb}_t$. Specifically, we design an interaction consistency reward that measures the interaction difference between the new agent and the demonstration:
\begin{align}
\interconsistR_t \propto -d(\hat{\IG}^{emb}_t, \IG^{emb}_t).
\end{align}
Here, $d$ indicates a distance metrics, the lesser the difference the larger the reward. The details measurement will be elaborated in the later parts. We use this reward to train a character control policy in simulation to learn the interaction movements of the new characters using reinforcement learning. In this subsection, we first elaboration how we build the crossponding interaction graph on new character setting, then we introduce the design of the interaction consistency reward, followed by some system design details.

\subsubsection{Formulating the Corresponding Embedded Interaction Graph}
Because the new agent may have a different morphology from the demonstration characters, we need to build a  sparse interaction graph on new characters before measuring the interaction consistancy. Specifically, we need to find the correspondence between the vertices of the embedded interaction graph $V^{emb}_t$ and the new character's vertices $\hat{V}^{emb}_t$, assuming that the edge relations remain the same.

While manual assignment of vertices is feasible due to the limited number of edges and vertices in the embedded graph, we opt for an automated vertices assigning mechanism based on positional alignment~\cite{li2023ace}. We start with assigning the end-effector vertices of the demonstrator humanoid character to the new characters' end-effectors. Given both characters' netural poses, for each end-effector of the character, we compute a vector that points from the root joint to the end-effector joint. We then compute the inner product between this vector and the vector from the root joint to each end-effector of the new character. The end-effector with the highest inner product is assigned as the corresponding end-effector of the new character. This process is repeated for all end-effectors, ensuring that each end-effector in the new character is assigned to a unique end-effector in the embedded graph. For the remaining vertices, we assign them according to their relative sequence to the end-effectors. So the vertices next to the end-effectors are assigned to the corresponding vertices in the new character. This process continues until all vertices in the embedded graph are assigned to the new character's vertices.

\subsubsection{Interaction Imitation Reward}
Once we adapt the embedded graph to the new character settings, we employ it to measure the graph distance using three metrics: length metrics $d^l$, root edge direction metrics $d^{ed}$, and center point metrics $d^{cp}$.

The length metrics $d^l$ assess the normalized length differences between the edges in the embedded graph of the new setting and the referenced graph:
\begin{align}
    d^l_t = \sum^{k-1}_{m=0} \left|\frac{l(\hat{\phi}^{m}_t)}{\hat{L}} - \frac{l({\phi}^{m}_t)}{{L}}\right|.
\end{align}
Here, $k$ indicates the total number of edges in the referenced embedded graph at time instance $t$. $l(\hat{\phi}^{m}_t)$ and $l({\phi}^{m}_t)$ represent the length of the $m$-th edge in the current and referenced graphs, respectively. $\hat{L}$ and ${L}$ are morphology-dependent length values used for normalization which are predefined parameters defined according to the morphology of the agent.

The root edge direction metric $d^{ed}$ measures the alignment of the root connection edge’s direction in the XY-plane. Here, the root connection edge refers to the edge that links the root vertices of the two characters. The objective is to ensure that the heading direction of this edge is aligned between the demonstration and the generated motion. The metrics is measured as:
\begin{align}
    d^{ed}_t = \frac{\langle(\hat{\phi}^{0}_t)_{xy}, ({\phi}^{0}_t)_{xy}\rangle}{l(\hat{\phi}^{0}_t) \cdot l({\phi}^{0}_t)}.
\end{align}
Here, $(\hat{\phi}^{0}_t)_{xy}$ denotes the roots connection edge vector projected onto the XY plane. 

The center point metrics $d^{cp}$ aim to align the height of the center point of each edge to a reference height:
\begin{align}
    d^{cp}_t = \sum^{M-1}_{m=0} \left|\frac{h(\hat{\phi}^{m}_t)}{\hat{L}} - \frac{h({\phi}^{m}_t)}{L}\right|.
\end{align}
Here, $h(\hat{\phi}^{m}_t)$ and $h({\phi}^{m}_t)$ denote the center point height of the $m$-th edge in the current and referenced graphs, respectively.

We employ a multiplicative approach to calculate the interaction consistency reward, addressing all components of the reward simultaneously. To mitigate the issue of low reward values when the current state deviates significantly from the reference, we incorporate an additional term that solely considers the length metric $d^{far}_t = d^l_t$ to handle large discrepancies. This low reward value issue typically occurs at the initial stages of the interaction. The interaction consistency reward is formulated as follows:
\begin{align}
    \interconsistR_t = 0.9 \exp(-w^{l}d^l_t)  \exp(-w^{ed}d^{ed}_t)  \exp(-w^{cp}d^{cp}_t) + 0.1  \exp(-w^{far}d^{far}_t).
\end{align}
Here, $w^{l}, w^{ed}, w^{cp},$ and $w^{far}$ represent the weights assigned to each component of the reward. The weight $w^{far}$ is set significantly lower than $w^{l}$, $w^{ed}$, and $w^{cp}$ to prevent the overall reward value from becoming too small, which can impair the learning algorithm's effectiveness.

\subsubsection{Regularization Rewards}
In addition to the interaction consistency reward, we incorporate several regularization rewards to improve the physical plausibility and stability of the learned motions. Following CrossLoco~\citep{li2023crossloco}, we include root height regularization, lateral orientation regularization, and torque regularization. These terms encourage the agent to maintain a consistent root height, minimize excessive lateral rotations, and avoid generating unrealistic joint torques during interaction. Together, these regularizers help ensure that the resulting behaviors are smooth, stable, and physically viable across diverse character embodiments.

\subsubsection{Centralized Hierarchical Policy Architecture}

To support multi-agent interaction control, we adopt a centralized hierarchical policy architecture. The system consists of a single high-level policy and multiple low-level sub-policies. The high-level policy coordinates overall interaction dynamics by aggregating observations from all characters and producing a latent control vector. Each low-level policy then takes this latent vector—along with the local sensory inputs of its assigned character—as input to produce motor commands. This hierarchical design enables scalable control across multiple agents while facilitating coordinated interaction behaviors. When characters share identical morphology, the corresponding low-level sub-policy is shared to promote parameter efficiency and behavioral consistency.


We pretrain each character's low-level policy using motion primitive pretraining, inspired by Imitation-and-Repurpose~\citep{bohez2022imitate}. During pretraining, each low-level controller learns to imitate motion distributions from a character-specific dataset. A high-level imitation policy provides latent control vectors based on privileged motion information, which is discarded after pretraining. This procedure enables the low-level policy to learn generalizable motion dynamics independently of specific reference trajectories. The resulting motion primitives serve as a strong initialization for downstream interaction learning. While the high-level policy is trained from scratch during interaction learning, the low-level controller continues to be fine-tuned, allowing it to adapt to interaction-specific objectives while retaining prior motor knowledge.